\documentclass[preprint,10.5pt]{elsarticle}

\usepackage[utf8]{inputenc} 
\usepackage[T1]{fontenc}    
\usepackage{hyperref}       
\usepackage{url}            
\usepackage{booktabs}       
\usepackage{amsfonts}       
\usepackage{nicefrac}       
\usepackage{microtype}      
\usepackage{lipsum}
\usepackage{graphicx}       
\graphicspath{{media/}}     
\usepackage{amsmath, amssymb}
\usepackage{float}
\usepackage{palatino}
\usepackage{verbatim}
\usepackage{mathrsfs}
\usepackage{array,delarray}
\usepackage{subcaption}
\usepackage{latexsym}
\usepackage{amscd}
\usepackage{xcolor}
\usepackage{mathtools}
\usepackage{array}

\usepackage{amsthm}
\usepackage{algorithm} 
\usepackage{algpseudocode} 
\usepackage{multirow}

\newcommand{\RR}{\mathbb{R}}

\newtheorem{theorem}{Theorem}

\newtheorem{proposition}{Proposition}

\newtheorem{remark}[theorem]{Remark}

\newtheorem{definition}[theorem]{Definition}

\begin{document}

\begin{frontmatter}

\title{Physics-Modeled Neural Networks}


\author[inst1]{Raul Felipe-Sosa}
\ead{rfelipe@nebrija.es}

\author[inst2]{Angel Martin del Rey}
\ead{delrey@usal.es}

\author[inst3]{Maria Flores Ceballos\corref{cor1}}
\ead{mariafc2552@usal.es}

\cortext[cor1]{Corresponding author.}

\address[inst1]{Departamento de Matemáticas, Escuela Politénica Superior, Universidad Nebrija, 28015 Madrid, Spain}

\address[inst2]{Department of Applied Mathematics, IUFFyM, Universidad de Salamanca, 37008 Salamanca, Spain}

\address[inst3]{Unidad de Excelencia en Luz y Materia Estructuradas (LUMES), Universidad de Salamanca, 37008 Salamanca, Spain}

\begin{abstract}
In this work, we introduce a novel continuous-time deep learning architecture, termed \emph{Dynamical Physics-Modeled Neural Networks} (DynPMNNs). Its main characteristic is that each hidden layer is defined as the solution of an ordinary differential equation, whose state variables represent the internal dynamics of the layer. This formulation generalizes classical feed-forward neural networks by replacing static activation functions with time-evolving dynamical systems, thereby providing a biologically inspired interpretation of hidden-layer behavior and enabling the direct incorporation of physical models into the network architecture. Our construction is rigorously grounded in the framework of Reproducing Kernel Banach Spaces (RKBSs), which allows us to characterize DynPMNNs as finite-dimensional solutions of an abstract training problem and reveals deep structural connections with classical neural networks.

We present a concrete implementation in which the hidden-layer dynamics follow the FitzHugh-Nagumo model, a reduced biophysical system describing neuronal activation. The resulting architecture integrates numerical ODE solvers, implemented via Euler-type blocks, into the computational graph and jointly trains both network weights and dynamical parameters. Through a regression task on the California Housing dataset, we empirically compare DynPMNNs with Neural Ordinary Differential Equations (NODEs) and Closed-form Continuous-Time Networks (CfCs). Despite having substantially fewer trainable parameters, the proposed model achieves competitive performance, illustrating the expressive power and efficiency that arise from embedding physically meaningful dynamics into neural architectures.

These results highlight the potential of DynPMNNs as a principled mathematical bridge between dynamical systems theory and deep learning, opening avenues for future work on expressivity, stability, numerical analysis, and physics-based learning.
\end{abstract}


\end{frontmatter}

\section{Introduction}

The integration of differential equations into deep learning architectures has led to a renewed perspective on neural networks as dynamical systems. Since the foundational works of McCulloch and Pitts \cite{McCulloch-1943} and Rosenblatt \cite{Rosenblatt-1958}, neural models have evolved from biologically inspired logical units into highly expressive computational structures capable of approximating complex functions. In recent years, this evolution has increasingly shifted toward continuous-time formulations, motivated by the desire to model temporal dynamics more faithfully and to exploit the analytical tools provided by differential equations. Two particularly influential approaches in this direction are Neural Ordinary Differential Equations (NODEs) \cite{chen-2018,kidger-2021} and Closed-form Continuous-Time Neural Networks (CfCs) \cite{hasani2022closed}, which, broadly speaking, reinterpret layer-to-layer propagation as the numerical integration of an initial value problem.

A key conceptual insight underlying NODEs is that a fully connected neural network can be reinterpreted by allowing each hidden layer to possess a continuously evolving internal state. Rather than viewing the activation function as a fixed nonlinear mapping, it may be regarded as a family of transformations parameterized by a continuous variable $t$, corresponding to the evolution of a dynamical system. From this perspective, classical feed-forward architectures can be interpreted as time discretizations of ordinary differential equations, and NODEs arise by taking the continuous-time limit of this process. This viewpoint generalizes classical feed-forward networks, in which activations are static, and provides a convenient mathematical framework for introducing temporal continuity into deep learning models.

However, within the NODE framework, differential equations primarily serve as a formal mechanism for introducing continuous-time dynamics. A NODE can be viewed as an ordinary differential equation whose right-hand side is represented by a neural network, without any intrinsic physical, biological, or physiological interpretation of either the dynamics or the trainable parameters involved. In particular, the state variables of a NODE correspond to abstract feature representations evolving in time, rather than to physically meaningful quantities. As a result, while NODEs provide a powerful tool for modeling temporal evolution, they do not explicitly aim to endow the internal dynamics of neural networks with mechanistic or interpretable structure.

In parallel, the incorporation of physical principles into machine learning models has attracted growing attention. In this context, Physics-Informed Neural Networks (PINNs) \cite{Karniadakis2019} embed the governing equations of a physical system directly into the loss function, thereby improving data efficiency and enforcing physically consistent behavior. However, PINNs do not modify the underlying neural architecture; rather, they impose physical constraints solely at the level of the output. By contrast, alternative approaches seek to integrate physical knowledge directly into the architecture itself, modeling neuron- or layer-level dynamics through systems of ordinary differential equations inspired by biophysical or mechanistic processes. Liquid Neural Networks (LNNs) \cite{hasani2021liquid}, for instance, represent a significant step in this direction by introducing adaptive continuous-time dynamics motivated by biological neural circuits.

In this work, we develop a mathematically rigorous framework for a new class of architectures, which we term \emph{Dynamical Physics-Modeled Neural Networks} (DynPMNNs). Our approach departs fundamentally from NODE-based formulations by explicitly modeling each hidden layer as the solution of an ODE system chosen to represent a meaningful physical, biological, or physiological process. In this setting, each neuron in a hidden layer is endowed with a well-defined internal state that evolves in time according to prescribed governing equations. As a consequence, the internal variables and trainable parameters of the network acquire a concrete mechanistic interpretation, rather than serving as purely abstract quantities.

To formalize this construction, we rely on the theory of Reproducing Kernel Banach Spaces (RKBSs) \cite{bartolucci-2023}, which provides a functional-analytic framework for deriving neural architectures as finite-dimensional solutions of abstract training problems. This perspective ensures that the proposed models remain mathematically consistent with the interpretation of neural networks as universal approximation tools while substantially generalizing classical feed-forward structures by embedding structured dynamical systems at the level of individual layers.

The DynPMNN framework allows the physical nature of each layer to be specified through its governing ODE system. As a result, hidden layers may encode biophysical dynamics (e.g., Hodgkin--Huxley or FitzHugh--Nagumo models), epidemiological processes, chemical reaction kinetics, or any system whose evolution is described by differential equations. This flexibility enables the direct incorporation of prior scientific knowledge into the network architecture. From a dynamical viewpoint, information propagates across layers along trajectories determined by the ODE flow, and the choice of the governing dynamics plays a central role in shaping the expressive power and qualitative behavior of the network.

To illustrate the proposed framework, we develop a concrete implementation in which the hidden-layer dynamics are governed by the FitzHugh--Nagumo neuronal model, a reduced two-dimensional variant of the Hodgkin--Huxley equations that captures essential features of neuronal excitation. The resulting DynPMNN is trained using a numerical Euler-type integration block embedded within the computational graph and evaluated on a regression task, where its performance is compared with NODE and CfC models. Despite involving a relatively small number of trainable parameters, the proposed architecture achieves competitive performance, highlighting the potential of physically structured neural models for efficient and interpretable learning.

Finally, beyond the introduction of a new neural architecture, the DynPMNN framework can be interpreted from the perspective of mathematical modeling as a novel methodology for the systematic integration of data-driven models with dynamical systems governed by ordinary differential equations. In this sense, the learning process may be viewed as a mechanism through which ODE-based models are informed and calibrated using data, giving rise to what may be described as \emph{data-informed ordinary differential equations}. This viewpoint establishes a new paradigm for coupling machine learning and mathematical modeling, in which data and structured dynamical laws interact coherently within a unified analytical framework.

The remainder of this article is organized as follows. Section 2 reviews the main approaches for integrating differential equations into neural network models, with emphasis on continuous-time and physics-informed architectures. Section 3 introduces the framework of Reproducing Kernel Banach Spaces (RKBSs) and its connection with classical fully connected neural networks, which provides the theoretical foundation for the proposed construction.

In Section 4, we define Dynamical Physics-Modeled Neural Networks (DynPMNNs), in which each hidden layer is modeled as the solution operator of an ordinary differential equation. This formulation yields a mathematically consistent and interpretable architecture that embeds mechanistic, physics- or physiology-inspired dynamics directly into the network. A concrete realization based on the FitzHugh–Nagumo neuronal model and its numerical implementation via embedded Euler integration blocks is presented.

Section 5 reports numerical experiments on a regression benchmark, where DynPMNNs are compared with Neural Ordinary Differential Equations (NODEs) and Closed-form Continuous-time Networks (CfCs). Despite using a remarkably small number of trainable parameters, DynPMNNs achieve competitive predictive performance, highlighting a favorable trade-off between model compactness, interpretability, and computational cost.

Finally, Section 6 summarizes the main conclusions and discusses future directions. Beyond proposing an efficient continuous-time neural architecture, this work positions DynPMNNs as a modeling paradigm for describing how incoming information can activate internal dynamical processes of a physiological nature, thereby bridging data-driven learning and differential-equation-based models of information processing.

\section{The study of physical systems using neural networks}

Over the past decades, theoretical studies on the expressivity of neural networks have established their ability to approximate complex functions. However, the development of architectures capable of efficiently capturing the underlying dynamics of physical and biological systems has motivated growing interest in incorporating mathematical tools, such as differential equations, into neural network design. This line of research has led to the emergence of new architectures, including Neural Ordinary Differential Equations (NODEs) and Physics-Informed Neural Networks (PINNs), which employ ordinary and partial differential equations, respectively \cite{chen-2018,Karniadakis2019}.

Differential equations play a central role in mathematical modeling, as they describe temporal or spatio-temporal variations in the state variables of a system. Their incorporation into neural network frameworks allows models not only to learn relationships from data but also to encode the physical laws governing the dynamics of the system under study. This approach has been shown to improve predictive accuracy and generalization while simultaneously reducing the need for large training datasets, since the model can infer system behavior based on underlying physical principles \cite{Karniadakis2021}.

The use of ordinary differential equations (ODEs) in the design of \emph{continuous-time} neural networks, such as NODEs, transforms the forward propagation process into a continuous integration problem, in which time is treated as a continuous variable \cite{chen-2018}. This formulation facilitates the modeling of continuous-time systems and enables the prediction of complex trajectories within a principled dynamical framework.

More specifically, NODEs are neural networks in which the evolution of latent states is governed by an ordinary differential equation. Instead of representing forward propagation as a discrete composition of layers, NODEs rely on a continuous-time model defined by the following initial value problem:
\begin{eqnarray} 
&&\frac{d\mathbf{x}(t)}{dt} = f\bigl(\mathbf{x}(t),t,\boldsymbol{\theta}\bigr), \label{eq:node}\\ 
&&\mathbf{x}(0)=\mathbf{x}_0,\quad t\in[0,T], \nonumber
\end{eqnarray}
where $\mathbf{x}(t)$ denotes the latent state at time $t$, and $f\bigl(\mathbf{x}(t),t,\boldsymbol{\theta}\bigr)$ is a neural network parameterized by $\boldsymbol{\theta}$, which defines the vector field of the dynamical system.

Inference in NODEs consists of numerically solving the initial value problem \eqref{eq:node} to obtain the terminal state $\mathbf{x}(T)$. This state is then mapped through an output layer to produce the final prediction. Training is carried out by minimizing a loss function $\mathcal{L}$ with respect to the parameters $\boldsymbol{\theta}$. To efficiently compute gradients through the ODE solution, the adjoint sensitivity method is commonly employed.

On the other hand, Physics-Informed Neural Networks (PINNs) constitute a different paradigm for incorporating physical knowledge into machine learning models. In PINNs, the governing equations of the system under study are explicitly embedded in the learning process through the loss function, which accounts for both observational data and the underlying physical laws \cite{Karniadakis2019}. As a result, PINNs can be used to approximate the solutions of highly nonlinear partial differential equations \cite{Karniadakis2021}, a task that is often computationally demanding when approached using traditional numerical methods.

Let $\mathcal{P}(\mathbf{u},\mathbf{x},t)=0$ denote the differential equation describing the system, where $\mathcal{P}$ is a (possibly nonlinear) differential operator and $\mathbf{u}=\mathbf{u}(\mathbf{x},t)$ is the unknown solution. A PINN approximates $\mathbf{u}(\mathbf{x},t)$ by means of a neural network $\hat{\mathbf{u}}(\mathbf{x},t;\boldsymbol{\theta})$, parameterized by $\boldsymbol{\theta}$. To enforce physical consistency, the loss function is defined as
\[
\mathcal{L} = \mathcal{L}_{\text{data}} + \mathcal{L}_{\text{physics}},
\]
where $\mathcal{L}_{\text{data}}$ measures the discrepancy between the available data $\mathcal{D}_d=\{\mathbf{x}_i,t_i,\mathbf{u}_i\}_{i=1}^{N_d}$ and the network prediction,
\begin{equation}
\mathcal{L}_{\text{data}} = \frac{1}{N_d} \sum_{i=1}^{N_d} 
\left\| \hat{\mathbf{u}}(\mathbf{x}_i,t_i;\boldsymbol{\theta}) - \mathbf{u}_i \right\|^2,
\end{equation}
and $\mathcal{L}_{\text{physics}}$ penalizes violations of the governing equations,
\begin{equation}
\mathcal{L}_{\text{physics}} = \frac{1}{N_p} \sum_{j=1}^{N_p} 
\left\| \mathcal{P}\bigl(\hat{\mathbf{u}}(\mathbf{x}_j,t_j;\boldsymbol{\theta}),\mathbf{x}_j,t_j\bigr) \right\|^2,
\end{equation}
evaluated at a set of collocation points $\mathcal{D}_p=\{\mathbf{x}_j,t_j\}_{j=1}^{N_p}$ within the domain of interest.

Training a PINN therefore amounts to minimizing the loss function $\mathcal{L}$ with respect to $\boldsymbol{\theta}$, ensuring that the learned approximation is consistent with both the observed data and the governing physical laws.


\section{Artificial Neural Networks and Reproducing Kernel Banach Spaces}\label{sec1}
Having provided a brief overview of the integration of artificial neural networks with physics-based and mathematically structured models, we now introduce the concept of \emph{Reproducing Kernel Banach Spaces} (RKBSs) \cite{bartolucci-2023,Zhang2009}. This framework has proven to be particularly suitable for understanding, from a rigorous mathematical perspective, the structure of neural network models and the theoretical foundations underlying their remarkable ability to fit data. As will become clear later, RKBSs exhibit a close relationship with fully connected neural networks. Moreover, the representation theorem discussed below provides a systematic recipe for constructing deep learning architectures that extend beyond those commonly used in practice.

\begin{definition}
Let $\mathcal{X}$ be an arbitrary set and let $\mathcal{B}$ be a Banach space of functions $f:\mathcal{X}\rightarrow\mathbb{R}$, endowed with pointwise addition and scalar multiplication. We say that $\mathcal{B}$ is a \emph{Reproducing Kernel Banach Space} (RKBS) if, for every $x\in\mathcal{X}$, there exists a constant $C_x>0$ such that
\[
|f(x)| \leq C_x \|f\|_{\mathcal{B}}, \qquad \forall\, f\in\mathcal{B}.
\]
\end{definition}

This definition implies that, for every $x\in\mathcal{X}$, the point evaluation functional $ev_x:f\mapsto f(x)$ belongs to the dual space $\mathcal{B}'$.

RKBSs can be seen as a natural generalization of reproducing kernel Hilbert spaces (RKHSs). In the Hilbert setting, one can identify a kernel function $K(x,y)$ such that $K(x,\cdot)\in\mathcal{B}$ for all $x\in\mathcal{X}$ and
\[
f(x)=\langle K(x,\cdot),f\rangle,
\]
where $\langle\cdot,\cdot\rangle$ denotes the inner product in $\mathcal{B}$. In the Banach setting, however, $\mathcal{B}$ cannot, in general, be identified with its dual $\mathcal{B}'$, and the existence of a reproducing kernel is no longer guaranteed. Nevertheless, a suitable reproducing property can still be established through the notions of \emph{characteristic space} and \emph{characteristic map}. The following proposition, proved in \cite{bartolucci-2023}, makes this precise.

\begin{proposition}\label{prop1}
A Banach space $\mathcal{B}$ of functions $f:\mathcal{X}\rightarrow\mathbb{R}$ is an RKBS if and only if there exist a Banach space $\mathcal{F}$ and a mapping $\phi:\mathcal{X}\rightarrow\mathcal{F}'$ such that:
\begin{itemize}
\item[(i)] $\mathcal{B}=\{f_{\mu}:\mu\in\mathcal{F}\}$, where
\[
f_{\mu}(x)=\prescript{}{\mathcal{F}'}{\langle\phi(x),\mu\rangle}_{\mathcal{F}};
\]
\item[(ii)] 
\[
\|f\|_{\mathcal{B}}=\inf\{\|\mu\|_{\mathcal{F}}: f=f_{\mu}\}.
\]
\end{itemize}
\end{proposition}

The space $\mathcal{F}$ is referred to as the \emph{characteristic space}, while the mapping $\phi$ is called the \emph{characteristic map}. Proposition~\ref{prop1} provides a constructive procedure for defining RKBSs once $\mathcal{F}$ and $\phi$ are specified. In what follows, we describe—without explicitly stating a separate theorem—a result whose proof is given in \cite{bartolucci-2023}, and which relies on a more general framework developed in \cite{Bredies2020}. This result establishes a remarkable connection between RKBSs and fully connected neural networks with a single hidden layer.

We begin by constructing an RKBS using Proposition~\ref{prop1}. To this end, we define $\mathcal{F}=\mathcal{M}(\Theta)$, the space of bounded Radon measures on the Borel $\sigma$-algebra of a parameter set $\Theta$. Endowed with the total variation norm $\|\cdot\|_{TV}$, the space $\mathcal{M}(\Theta)$ is a Banach space that can be identified with the dual of $C_0(\Theta)$, the space of continuous functions on $\Theta$ vanishing at infinity. In this setting, the norm $\|\mu\|_{TV}$ coincides with the operator norm of $\mu$ as an element of $C_0(\Theta)'$.

Next, we define the characteristic map. Let
\begin{itemize}
\item $\rho:\mathcal{X}\times\Theta\rightarrow\mathbb{R}$ be such that, for every $x\in\mathcal{X}$, the function $\rho(x,\cdot)$ is measurable;
\item $\beta:\Theta\rightarrow\mathbb{R}$ be such that $\rho(x,\cdot)\beta(\cdot)\in C_0(\Theta)$.
\end{itemize}
The characteristic map $\phi:\mathcal{X}\rightarrow\mathcal{M}(\Theta)'$ is then defined by
\begin{align}\label{eqn1}
\prescript{}{\mathcal{M}(\Theta)'}{\langle\phi(x),\mu\rangle}_{\mathcal{M}(\Theta)}
=\int_{\Theta}\rho(x,\theta)\beta(\theta)\,d\mu(\theta).
\end{align}
The assumptions on $\rho$ and $\beta$ ensure that the integral in \eqref{eqn1} is well defined and finite, and they also guarantee additional properties that will be relevant for the main results discussed below.

An RKBS $\mathcal{B}$ can thus be defined as
\[
\mathcal{B}
=\bigl\{f_{\mu}:\mu\in\mathcal{M}(\Theta)\bigr\},\qquad
f_{\mu}(x)=\int_{\Theta}\rho(x,\theta)\beta(\theta)\,d\mu(\theta),
\]
with norm
\[
\|f\|_{\mathcal{B}}
=\inf\{\|\mu\|_{TV}: f=f_{\mu},\ \mu\in\mathcal{M}(\Theta)\}.
\]

Let $\{(x_i,y_i)\}_{i=1}^{N}\subset\mathcal{X}\times\mathbb{R}$ denote a finite training dataset. The objective is to determine a function $f\in\mathcal{B}$ that provides an optimal fit to the data. This objective can be rigorously formulated through the following abstract training problem:
\begin{align}\label{eqn2}
\inf_{f\in\mathcal{B}}
\left(
\frac{1}{N}\sum_{i=1}^{N}L\bigl(y_i,f(x_i)\bigr)
+\|f\|_{\mathcal{B}}
\right).
\end{align}

A fundamental result, proved in \cite{bartolucci-2023} (Theorem~3.9), states that the minimizers of \eqref{eqn2} belong to a finite-dimensional subspace of $\mathcal{B}$ of dimension $M\leq N$ and admit the representation
\begin{align}\label{eqn3}
f(x)
=\sum_{k=1}^{M}a_k
\prescript{}{\mathcal{M}(\Theta)'}{\langle\phi(x),\delta_{\theta_k}\rangle}_{\mathcal{M}(\Theta)}
=\sum_{k=1}^{M}\alpha_k\,\rho(x,\theta_k),
\end{align}
where $a_k\in\mathbb{R}\setminus\{0\}$, $\delta_{\theta_k}$ denotes the Dirac measure supported at $\theta_k\in\Theta$, and $\alpha_k=a_k\beta(\theta_k)$ with $\beta(\theta_k)\neq0$.

This result has important theoretical and practical implications. In particular, it shows that the infinite-dimensional optimization problem \eqref{eqn2} can be reduced to a finite-dimensional one, with dimension bounded by the number of training samples.

An additional and particularly significant consequence is the direct connection between RKBSs and fully connected neural networks with a single hidden layer. Indeed, by choosing $\mathcal{X}=\mathbb{R}^n$, $\Theta=\mathbb{R}^{n+1}$ with $\theta=(w,b)$, and defining $\rho(x,\theta)=\sigma(w\cdot x+b)$ for a given activation function $\sigma:\mathbb{R}\rightarrow\mathbb{R}$, the representation \eqref{eqn3} corresponds exactly to a fully connected neural network with one hidden layer of $M$ neurons. In this interpretation, each generator of the finite-dimensional subspace appearing in \eqref{eqn3} represents a neural unit in the hidden layer. More generally, these generators may be interpreted as fundamental structural units of the architecture under consideration.

The preceding analysis highlights that RKBSs provide a natural theoretical framework in which new deep learning architectures arise as finite-dimensional subspaces associated with solutions of abstract training problems. While the single-hidden-layer case serves as a canonical example, more general RKBS constructions satisfying the assumptions of Theorem~3.9 in \cite{bartolucci-2023} give rise to novel deep architectures beyond standard feed-forward networks.

In the following sections, we build upon this methodology to define neural networks whose hidden layers are composed of units governed by ordinary differential equations. These architectures, which we refer to as \emph{Physics-Modeled Neural Networks} (PMNNs), form the mathematical foundation upon which the dynamical models introduced later in this work are constructed.

\section{Dynamical Physics-Modeled Neural Networks}
Although there are many types of mathematical models, for the purposes of this work we distinguish two broad categories: \emph{physics-based models} and \emph{data-based models}. The former are typically formulated in terms of differential equations, either ordinary or partial, and can be viewed as mathematical translations of the fundamental laws governing physical phenomena. When only limited experimental observations are available, the theoretical analysis of such models can still yield substantial quantitative and qualitative insight into the phenomenon of interest. In this regime, one may say that there is \emph{much physics and little data}. Physics-based models support tasks such as forecasting future scenarios and analyzing causes that cannot be directly observed or measured. Thus, beyond predictive capabilities, these models provide high-quality qualitative understanding of the systems they describe.

At the other end of the spectrum lie data-based models, where abundant information is available in the form of large datasets, while the physics underlying the observations is not explicitly incorporated into the model. Such models are trained primarily to perform tasks such as prediction or classification. In the terminology above, this corresponds to the regime of \emph{much data and little (or no) physics}. We refer to Box~1 of \cite{Karniadakis2021} for a schematic summary of this viewpoint.

Between these two extremes, there is an intermediate regime in which one has \emph{some physics} and \emph{some data}. For instance, one may consider a system of differential equations with unknown terms (e.g., unknown sources or physical parameters), together with additional measurements that, beyond boundary or initial conditions, enable the solution of inverse problems to identify the missing quantities. More generally, there are multiple strategies for incorporating physics-based models into deep learning architectures, that is, into data-driven models. In \cite{Karniadakis2021}, these strategies are described in terms of biases, and three categories are discussed: observational biases, inductive biases, and learning biases. Since our interest here is specifically in integrating physics into data-based models through differential-equation structure, we summarize the discussion into the following two approaches.

\paragraph{Approach 1.}
This line of research focuses on constructing deep architectures by modeling processes involved in brain activity, such as the electrochemical activation of neurons and the propagation of activation to neighboring cells. In this setting, cell-activation models expressed as ordinary differential equations (ODEs) are explicitly incorporated into the architecture.

Models developed under this approach can process data of various types and are typically used for prediction and classification. It is natural to expect that embedding mechanistic models of neural processes may improve predictive performance, although the theoretical understanding of this improvement remains an open question. When the models integrated into the architecture are ODE systems, we refer to the resulting architectures as \textbf{\emph{Dynamical Physics-Modeled Neural Networks}} (DynPMNNs). These architectures constitute the main object of study in this work.

\paragraph{Approach 2.}
A second approach incorporates differential equations into neural networks by imposing the constraint that the \emph{output} of the network must satisfy a differential equation, typically by including the governing equations explicitly in the loss function. In this case, the network serves two purposes: approximating the solution of the differential equation and addressing inverse problems associated with it. Physics-Informed Neural Networks (PINNs) provide a representative example of this paradigm; see \cite{Karniadakis2019}.

As mentioned earlier, the goal of this article is to construct a new neural architecture by assuming that each hidden layer can be modeled—under assumptions specified below—by a system of ODEs. We propose to do this in two ways, which we describe next.

\subsection{PMNNs with dynamical neural units: DynPMNNs}

We construct a new deep architecture following the methodology outlined in Section~\ref{sec1}. To this end, we first build an appropriate RKBS.

Let us consider the following initial value problem.
\begin{align}\label{PVI}
\left\{
\begin{array}{l}
\dfrac{dx}{dt} = g(x, \tilde{\theta}),\\
x(0) = x_0,
\end{array}
\right.
\end{align}
where $g: \RR^{m} \times \RR^{p} \rightarrow \RR^{m}$ is such that, for all $x_0 \in \RR^m$ and $\tilde{\theta} \in \RR^p$, this initial value problem admits a solution defined for all $t > 0$. In this sense, we can define a one-parameter family of operators $\sigma^{(t)}: \RR^{m}\times \RR^p \rightarrow \RR^m$ representing the flow of a dynamical system in the variables $t$ and $x$. That is, for fixed $\tilde{\theta} \in \RR^p$ and $t > 0$, $\sigma^{(t)}(x_0; \tilde{\theta})$ represents the state obtained by evolving $x_0$ up to time $t$.

Now, let $\mathcal{X} = \RR^n$ and $\Theta = [0, +\infty)\times \RR^n\times \RR\times \RR^p$. As in Section~\ref{sec1}, we define $\mathcal{F} = \mathcal{M}(\Theta)$ with the total variation norm, $\|\cdot\|_{TV}$. The characteristic map is defined as follows:
\begin{eqnarray*}
\rho: \mathcal{X}\times \Theta &\rightarrow& \RR^m\\
(x, t, w, b, \tilde{\theta})&\mapsto&\rho(x, t, w, b, \tilde{\theta}) = \sigma^{(t)}(w\cdot x + b; \tilde{\theta})
\end{eqnarray*}
where $w\cdot x$ represents the dot product in $\RR^n$. On the other hand, we define a function $\beta: \Theta \rightarrow \RR^m$ such that, for all $x \in \mathcal{X}$,
\[
\rho(x, \cdot)\cdot \beta \in C_{0}(\Theta),
\]
where $\cdot$ denotes the dot product in $\RR^{m}$.

We define $\phi: \mathcal{X} \rightarrow \mathcal{M}(\Theta)'$ such that
\[
\prescript{}{\mathcal{M}(\Theta)'}{\left\langle \phi(x), \mu\right\rangle}_{\mathcal{M}(\Theta)}
= \int_{\Theta}\sigma^{(t)}(w\cdot x + b, \tilde{\theta})\cdot \beta(\theta)\,d\mu(\theta).
\]

As a consequence, our RKBS is given by:
\begin{align*}
\mathcal{B}_{\text{DynPMNN}} &= \left\{f_{\mu}: \mu \in \mathcal{M}(\Theta)\right\},\\[0.15cm]
f_{\mu}(x) &=  \int_{\Theta}\sigma^{(t)}(w\cdot x + b, \tilde{\theta})\cdot \beta(\theta)\,d\mu(\theta),\\[0.15cm]
\|f\|_{\mathcal{B}_{\text{DynPMNN}}} &= \inf\left\{\|\mu\|_{TV}: \mu \in \mathcal{M}(\Theta),\, f = f_{\mu}\right\}. 
\end{align*}
We refer to $\mathcal{B}_{\text{DynPMNN}}$ as an \emph{abstract DynPMNN}.

Although our RKBS is somewhat different from the one originally given in \cite{bartolucci-2023}, it is straightforward to verify that Theorem 3.9 (as stated in \cite{bartolucci-2023}) applies to the abstract training problem associated with $\mathcal{B}_{\text{DynPMNN}}$. In this sense, we can say that a solution to this problem can be found in a finite-dimensional subspace, which we denote by $\mathcal{DN}^{(w, b)}_{t,\tilde{\theta}}$, consisting of functions of the form:
\begin{align*}
f(x) = \sum^{M}_{k = 1}\alpha_k \cdot \sigma^{(t_k)}(w_k\cdot x + b_k, \tilde{\theta}_k),
\end{align*}
where $\alpha_k \in \RR^m$, $w_k \in \RR^n$, $b_k \in \RR$, and $\tilde{\theta}_k$ for $k = 1,\ldots,M$ are trainable parameters, and $\{t_k\}_{k=1}^{M}$ is a set of points in the interval $(0, +\infty)$.

From now on, we will refer to the elements $\sigma^{(t_k)}(w_k\cdot x + b_k, \tilde{\theta}_k) \in \RR^m$ as \emph{DynPMNN neural units}. Thus, each generator of the subspace $\mathcal{DN}^{(w, b)}_{t,\tilde{\theta}}$ is one such unit. Structurally, these units can be represented as blocks whose dynamics are modeled by $m$ state variables, namely the components of the vector $\sigma^{(t_k)}(w_k\cdot x + b_k, \tilde{\theta}_k)$.

We next describe how these DynPMNN neural units appear when organized into a standard feed-forward architecture. Consider a fully connected neural network with an input layer, a hidden layer, and an output layer. We denote by $x_{\text{in}} \in \RR^{d_{\text{in}}}$ the input data vector of dimension $d_{\text{in}}$ (the number of predictor variables). We also consider $W_h x_{\text{in}} + b_h$ as the result of linearly connecting the input layer with the hidden layer, where $W_h \in \RR^{d_h\times d_{\text{in}}}$ is the weight matrix of the hidden layer, $b_h \in \RR^{d_h}$ is the bias vector, and $d_h$ is the number of neurons in the hidden layer.

We now introduce a one-parameter family of activation functions $x^{(t)}_h = \sigma^{(t)}_{h}(0, W_h x_{\text{in}} + b_{h})$ that determines the output of the hidden layer. Here $\sigma_{h}^{(t)}$ is the solution of the following initial value problem (IVP):
\begin{align}\label{ODE1}
\left\{
\begin{array}{c}
\dfrac{dx_h}{dt} = g^{(h)}(x_h,\tilde{\theta}_h),\\[0.15cm]
x_h(0) = W_h x_{\text{in}} + b_{h},
\end{array}
\right.
\end{align}
where $g^{(h)}$ is a vector field defined on $\RR^{d_h}\times \RR^{p_h}$ with values in $\RR^{d_h}$. Here, $p_h$ is the dimension of the parameter space that determines the vector field $g^{(h)}$. We assume that $g^{(h)}$ satisfies the appropriate conditions for the solution to be defined on an interval $I_h = [0, T_h]$ independent of the trainable weights $W_h$, $b_h$, and the input data $x_{\text{in}}$. We define the state of the hidden layer at time $t$ as $x^{(t)}_h \in \RR^{d_{h}}$ for all $t \in I_h$. In particular, each component of the hidden-layer output models a state variable of the layer at time $t$.

Similarly, the output of the output layer is calculated as follows. Fix $t \in I_h$ and define
$x^{(t, s)}_{\text{out}}= \sigma^{(s)}_{\text{out}}\left(t, W_{\text{out}}x^{(t)}_h + b_{\text{out}}\right)$, where $\sigma^{(s)}_{\text{out}}$ is the solution of the following IVP:
\begin{align*}
\left\{
\begin{array}{c}
\dfrac{dx_{\text{out}}}{ds} = g^{(\text{out})}(x_{\text{out}},\tilde{\theta}_{\text{out}}),\\
x_{\text{out}}(t) = W_{\text{out}}x^{(t)}_h + b_{\text{out}}.
\end{array}
\right.
\end{align*}
Here, $W_{\text{out}} \in \mathbb{R}^{d_{\text{out}}\times d_h}$ is the weight matrix of the output layer, $b_{\text{out}} \in \mathbb{R}^{d_{\text{out}}}$ is the bias vector of the output layer, $d_{\text{out}}$ is the number of neurons in the output layer, and  $\tilde{\theta}_{\text{out}} \in \mathbb{R}^{p_{\text{out}}}$. Again, we define $x^{(t, s)}_{\text{out}}$ as the state of the output layer at time $s$ originating from time $t$, with $(t, s) \in I_h \times I_{\text{out}}$ and $t < s$, where $I_{\text{out}} = [t, T_{\text{out}})$ (with $T_{\text{out}}$ possibly infinite).

Finally, we can say that this neural network is modeled by a two-parameter family of functions
\begin{align*}
f^{(t, s)}_{\theta}: \mathbb{R}^{d_{\text{in}}} \rightarrow \mathbb{R}^{d_{\text{out}}},\;\; \hbox{with $(t, s) \in I_h \times I_{\text{out}},$}
\end{align*}
such that $f^{(t, s)}_{\theta}(x_{\text{in}}) = \left(\mathrm{l}^{(s)}_{\text{out}}\circ \mathrm{l}^{(t)}_{h}\right)(x_{\text{in}})$, where:
\begin{align*}
\mathrm{l}^{(t)}_{h}(x_{\text{in}}) &= \sigma^{(t)}_h\left(W_h x_{\text{in}} + b_h\right),\\[0.15cm]
\mathrm{l}^{(s)}_{\text{out}}(x^{(t)}_{h}) &= \sigma^{(s)}_{\text{out}}\left(W_{\text{out}}x^{(t)}_h + b_{\text{out}}\right).
\end{align*}
Here, $\theta \in \mathbb{R}^{p}$ represents the trainable parameters of the network, which include $(W_h, b_h, W_{\text{out}}, b_{\text{out}}, \tilde{\theta}_{h}, \tilde{\theta}_{\text{out}})$, that is,
\begin{align*}
p = n_{\text{in}}n_{h} + n_{\text{out}}n_{h} + d_h + d_{\text{out}} + p_h + p_{\text{out}}.
\end{align*}
Note that we treat the parameters of the ODE systems modeling the dynamics of each layer, $\tilde{\theta}_h$ and $\tilde{\theta}_{\text{out}}$, as trainable.

After introducing this definition, several aspects must be discussed to clarify the scope and interpretation of these models beyond their practical implementation. Concerning the state modeling of each layer, we note the following. In classical fully connected multi-layer neural networks, the number of neurons in each hidden layer is a hyperparameter chosen by the user. In the DynPMNN setting, the user must first choose the dynamical model (an ODE system) describing the layer dynamics, and the number of neurons in the layer is then identified with the dimension of that system. In particular, to obtain hidden layers with many neurons, one must consider models with many state variables. In contrast to classical neural networks, this approach makes it meaningful to discuss the \emph{physical nature} of a layer, determined by the phenomenon modeled by the chosen ODE system. For example, one may speak of physiological layers when the selected model describes a physiological process, such as the electrical activation of neurons or cells; similarly, one may consider epidemiological or demographic layers, among others.

This interpretation may not be physiologically realistic if the goal is to simulate brain activity in detail, but it is mathematically meaningful, especially when assessing model performance. Although the theoretical analysis remains open, it is reasonable to expect that richer hidden-layer dynamics can contribute to improved accuracy.

An important question concerns how to interpret these networks as a whole, an issue that remains somewhat unclear in the literature. Each layer (hidden and output) evolves according to its own ODE system. However, the network as a whole cannot, in general, be interpreted as a single dynamical system. Indeed, the function representing the network, $f^{(t, s)}_{\theta}$, depends on two parameters, and the input and output spaces typically have different dimensions. The superscript $(t,s)$ is an ordered pair, since by definition $t<s$. In this sense, $x_{\text{out}} = f^{(t, s)}_{\theta}(x_{\text{in}})$ can be interpreted as a signal that propagates through the hidden layer, exits it at time $t\in I_h$, and exits the output layer at time $s\in I_{\text{out}}$. Thus, the pair $(t,s)$ encodes the times at which information passes from one layer to the next. From this viewpoint, the phenomenon of information propagation across layers underlies the network mapping, while the internal dynamics of the layers remain, in a sense, decoupled.

This perspective also suggests further extensions. So far, we have assumed a single direction of signal propagation and have not incorporated any spatial arrangement of the neurons within each layer. However, the present framework naturally accommodates the possibility that each layer, and its coupling to adjacent layers, carries a two- or three-dimensional geometry that influences the overall behavior. Such extensions are closely related to neural architectures derived from partial differential equations (PDEs) and may be relevant for modeling connectivity between brain areas.

We now generalize this construction to the case of $N-1$ hidden layers. Let $k=1,\ldots,N,N+1$, where $k=1$ is identified with the input layer (subscript \textbf{in}) and $k=N+1$ with the output layer. The output of the first hidden layer (layer $k=2$) is computed as follows: $\mathrm{l}^{(t^{(2)})}_1: \RR^{n_1} \rightarrow \RR^{n_2}$ with $W_2 \in \mathbb{R}^{n_2 \times n_1}$ and $b_2 \in \mathbb{R}^{n_2}$. Again, we consider that the state of layer $k=2$ is
\[
x_2^{(t^{(2)})} = \mathrm{l}^{(t^{(2)})}_1(x_{\text{in}}) = \sigma^{(t^{(2)})}_2(W_2 x_{\text{in}} + b_2),
\]
which is the solution of the ODE system
\begin{align*}
\left\{
\begin{array}{c}
\dfrac{dx_2}{dt^{(2)}} = g^{(2)}(x_{2},\tilde{\theta}_{2}),\\[0.15cm]
x_{2}(0) = W_2 x_{\text{in}} + b_2,
\end{array}
\right.\;\; t^{(2)} \in (0,T_2),
\end{align*}
where $g^{(2)}: \mathbb{R}^{n_2}\times \mathbb{R}^{p_2} \rightarrow \mathbb{R}^{n_2}$. 

Let us go one step further. Set $t^{(2)} = t_2 \in (0,T_2]$ (note that we denote the time variable controlling the dynamics of the layer by a superscript, and a fixed value of this variable by a subscript). The output of layer $k=3$ is then computed as
\[
\mathrm{l}^{(t_2, t^{(3)})}_2(x^{(t_2)}_2) = \sigma^{(t^{(3)})}_3\left(W_3 x^{(t_2)}_2 + b_3\right),
\]
where $W_3 \in \mathbb{R}^{n_3\times n_2}$, $b_3 \in \mathbb{R}^{n_3}$, and the output is obtained as the solution of the IVP
\begin{align*}
\left\{
\begin{array}{c}
\dfrac{dx_3}{dt^{(3)}} = g^{(3)}(x_{3},\tilde{\theta}_{3}),\\
x_{3}(t_2) = W_3 x^{(t_2)}_2 + b_3,
\end{array}
\right.\;\; t^{(3)} \in (t_2,T_3).
\end{align*}
Here $g^{(3)}: \mathbb{R}^{n_3}\times \mathbb{R}^{p_3} \rightarrow \mathbb{R}^{n_3}$.

Now define the output of layer $k+1$. Fix $\left(t^{(2)}, t^{(3)},\ldots, t^{(k)}\right) = (t_2, t_3,\ldots, t_k)$, and compute
\[
\mathrm{l}^{(t_k, t^{(k+1)})}_k(x^{(t_k)}_k) = \sigma^{(t^{(k+1)})}_{k+1}\left(W_{k+1} x^{(t_k)}_k + b_{k+1}\right),
\]
where $W_{k+1} \in \mathbb{R}^{n_{k+1}\times n_k}$, $b_{k+1} \in \mathbb{R}^{n_{k+1}}$, and this output is obtained as the solution of the following IVP:
\begin{align*}
\left\{
\begin{array}{c}
\dfrac{dx_{k+1}}{dt^{({k+1})}} = g^{({k+1})}(x_{k+1},\tilde{\theta}_{k+1}),\\
x_{k+1}(t_k) = W_{k+1} x^{(t_k)}_k + b_{k+1},
\end{array}
\right.\;\; t^{(k+1)} \in (t_k,T_{k+1}).
\end{align*}
where $g^{(k+1)}: \mathbb{R}^{n_{k+1}}\times \mathbb{R}^{p_{k+1}} \rightarrow \mathbb{R}^{n_{k+1}}$.

Finally, fix $\left(t^{(2)}, t^{(3)},\ldots, t^{(\text{out})}\right) = (t_2, t_3,\ldots, t_{\text{out}})$, and define the mapping associated with the dynamical network by
\begin{align*}
&f^{(t_2, t_3,\ldots, t_{\text{out}})}_{\theta}: \mathbb{R}^{n_1} \rightarrow \mathbb{R}^{n_{\text{out}}},\\
&f^{(t_2, t_3,\ldots, t_{\text{out}})}_{\theta}(x_1)
= \left(\mathrm{l}_N^{(t_N, t_{\text{out}})}\circ \mathrm{l}_{N-1}^{(t_{N-1}, t_{N})}\circ\cdots \circ \mathrm{l}_1^{(t_2)}\right)(x_1).
\end{align*}
Note that we have fixed the parameters determining the times at which information exits one layer and enters the next; hence, these times may be interpreted as hyperparameters of the network.

\begin{remark}
From this point on, we obtain a well-defined deep architecture that generalizes the classical feed-forward multi-layer network and enables us to model information flow through layers governed by prescribed ODE systems. When formulated in this way, these architectures appear particularly well suited for identifying parameters in ODE systems from time-resolved data. While there is likely related literature in this direction, it may be intriguing to explore parameter-identification and inverse problems within the present framework, including potential applications to malware propagation. Another important aspect is to understand how these networks are trained and what computational or approximation advantages they may offer. Numerous theoretical questions arise. For instance, as mentioned earlier, one may aim to formulate the training problem as the minimization of a functional over a Banach or Hilbert space of reproducing kernels, with potentially far-reaching implications for the theoretical analysis of these architectures.
\end{remark}

\subsection{Training a PMNN Network}

Having defined the concept of a Physics-Modeled Neural Network (PMNN), we now describe the training procedure associated with this class of architectures. In order to compute the output of a hidden layer, it is necessary to solve the system of ordinary differential equations that governs the dynamics of that layer. There are two main strategies to address this task: one may either incorporate a classical numerical time-integration scheme for ODEs directly into the PMNN training process, or integrate a Physics-Informed Neural Network (PINN) to approximate the solution of the ODE system. In principle, both strategies allow a PMNN to be trained consistently within the proposed framework.

Nevertheless, in the present work we focus exclusively on the first strategy and rely on standard explicit time-integration methods to solve the ODE associated with the hidden layer during both training and inference. This choice is motivated by several considerations.

First, PINN-based solvers typically introduce an additional learning problem, since the neural surrogate employed to approximate the ODE solution must itself be trained to sufficient accuracy, either prior to or jointly with the optimization of the PMNN parameters. As a consequence, the overall training procedure becomes more complex and computationally demanding. Second, in the low-dimensional ODE setting considered in this work, classical numerical schemes—such as the explicit Euler method—are computationally efficient, straightforward to implement, and easy to tune. Moreover, they are sufficient to capture the relevant dynamics of the hidden layer without incurring the additional overhead associated with PINN-based approaches.

For these reasons, the remainder of this section focuses on the integration of a classical Euler-type scheme into the PMNN architecture. In particular, we detail how the resulting Euler block is embedded within the computational graph of the network and how its parameters are optimized jointly with the remaining trainable weights of the model.



\subsubsection{Training a PMNN using the Euler method}

\begin{figure}[H]
\centering
\includegraphics[width=0.5\linewidth]{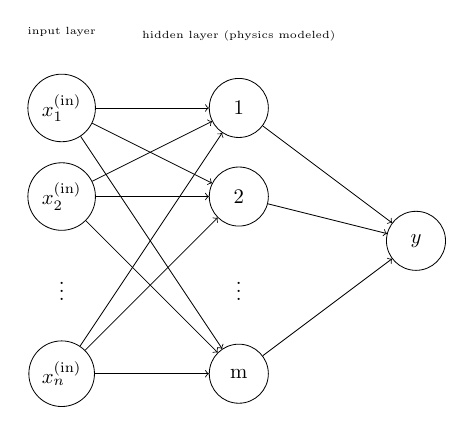}
\caption{Representation of an MLP with a single hidden layer.}
\label{fig:MLP1}
\end{figure}

We describe this training strategy by employing the Euler scheme to solve the initial value problem that models the hidden layer dynamics (see \cite{Butcher2016}). As a first illustrative example, we consider a PMNN with a single hidden layer governed by a system of ordinary differential equations. The input layer consists of $n$ neurons, while the hidden layer consists of $m$ neurons (see Figure~\ref{fig:MLP1}). We further assume that the activation function of the output layer is a classical activation function $\sigma^{(\text{out})}\colon \mathbb{R} \rightarrow \mathbb{R}$, as is commonly used in standard feed-forward architectures.

Recall that the output of the hidden layer is modeled by a system of ODEs. In this setting, our goal is to compute an approximate solution of an initial value problem similar to \eqref{ODE1} over a prescribed time interval $[0, T_h]$. To this end, we employ the explicit Euler method. We introduce $N+1$ time points in this interval, defined by
\[
t_0 = 0 < t_1 < t_2 < \cdots < t_N = T_h.
\]
According to the Euler method, an approximate solution of \eqref{ODE1} is given by
\begin{align*}
x_h(t) = x_{h}(t_{k-1}) + (t_k - t_{k-1})\, g^{(h)}\!\left(x_h(t_{k-1}), \tilde{\theta}_h\right),
\quad \forall\, t \in (t_{k-1}, t_k],
\end{align*}
for $k = 1,2,\ldots,N$. In particular, we are interested in computing the following sequence of approximations:
\begin{align}\label{EulerM}
\begin{cases}
x_{h}(t_1) = W^{(h)}x_{\text{in}} + b^{(h)} 
+ t_1\, g^{(h)}\!\left(W^{(h)}x_{\text{in}} + b^{(h)}, \tilde{\theta}_h\right),\\
x_{h}(t_2) = x_{h}(t_1) + (t_2 - t_1)\, g^{(h)}\!\left(x_{h}(t_1), \tilde{\theta}_h\right),\\
\hspace{0.7cm} \vdots  \\
x_{h}(t_k) = x_{h}(t_{k-1}) + (t_k - t_{k-1})\, g^{(h)}\!\left(x_{h}(t_{k-1}), \tilde{\theta}_h\right),
\end{cases}
\end{align}
where $k$ is chosen according to the desired final integration time. The output of the hidden layer is then defined as
\begin{align*}
\sigma^{(t_k)}_{h}\!\left(W^{(h)}x_{\text{in}} + b^{(h)}; \tilde{\theta}_h\right)
= x_h(t_k),
\end{align*}
with $W^{(h)} \in \mathbb{R}^{m\times n}$ and $b^{(h)} \in \mathbb{R}^m$ denoting the weight matrix and bias vector of the hidden layer, respectively.

Observe that the output of the hidden layer corresponds to the last term of the sequence in \eqref{EulerM}, and therefore depends explicitly on the weight matrix and bias of that layer. From an architectural point of view, incorporating the Euler method into the network effectively adds multiple intermediate layers, one for each step in the Euler recursion \eqref{EulerM}. These intermediate layers are referred to as \emph{Euler layers}, and their number constitutes a hyperparameter of the model. Increasing the number of Euler layers corresponds to advancing the hidden-layer state further in time relative to its initial condition.

The output of the $j$-th Euler layer is given by
\begin{align*}
x_{h}(t_j) = x_{h}(t_{j-1}) + (t_j - t_{j-1})\, g^{(h)}\!\left(x_{h}(t_{j-1}), \tilde{\theta}_h\right),
\end{align*}
where the term $x_{h}(t_{j-1})$ plays the role of a residual connection between consecutive Euler layers.

\begin{figure}
\centering
\includegraphics[width=0.8\linewidth]{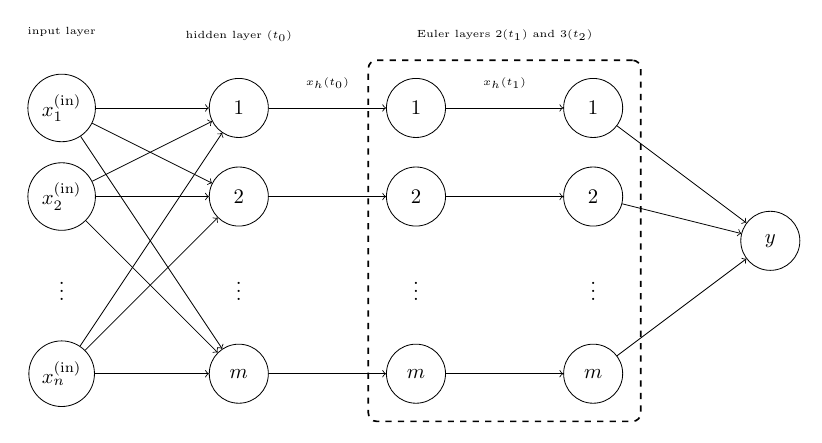}
\caption{Representation of a PMNN with a two-layer Euler block.}
\label{fig2}
\end{figure}

The collection of Euler layers will be referred to as the \emph{Euler block} (see Figure~\ref{fig2}).

Once the number of Euler layers has been selected, the output of the network—and hence the output of the entire PMNN—is given by
\begin{align*}
f_{(\theta, \tilde{\theta}_h)}(x_{\text{in}})
= \sigma^{(\text{out})}\!\left(W^{(\text{out})}x_h(t_k) + b^{(\text{out})}\right).
\end{align*}
Here, $W^{(\text{out})} \in \mathbb{R}^{1\times m}$ and $b^{(\text{out})} \in \mathbb{R}$ denote the weight matrix and bias of the output layer, respectively, and $\sigma^{(\text{out})}$ is the activation function applied at this layer. In this setting, the trainable parameters are
\[
\theta = \left(W^{(h)}, W^{(\text{out})}, b^{(h)}, b^{(\text{out})}\right)
\quad \text{and} \quad \tilde{\theta}_h,
\]
where $\tilde{\theta}_h$ corresponds to the parameters of the ODE system modeling the hidden layer dynamics.

We have thus defined a PMNN architecture in which an auxiliary Euler block is embedded to numerically solve the initial value problem governing the hidden layer. Within this framework, alternative numerical integration schemes may also be employed to solve the IVP, leading to different types of auxiliary blocks and potentially distinct architectural variants.

\subsubsection{FitzHugh--Nagumo Model for Modeling the Hidden Layer}

The Hodgkin--Huxley model (HH model) represents a major milestone in medical electrophysiology and mathematical modeling, as it successfully reproduces all the characteristic features of the neuronal action potential in a biologically realistic manner. Moreover, all the parameters of the model admit a clear physiological interpretation. However, the strong nonlinearity of the model and the dimensionality of its phase space—which is four-dimensional—significantly complicate both its qualitative analysis and the interpretation of the parametric regions associated with the different dynamical regimes that may arise.

As a consequence of these limitations, shortly after the formulation of the HH model, mathematicians became interested in deriving simplified models that preserve its essential dynamical behavior while reducing analytical complexity. Owing to the seminal contributions of FitzHugh and Nagumo, a reduced model was obtained that captures the fundamental mechanisms underlying the neuronal action potential while being considerably simpler than the original HH system.

The FitzHugh--Nagumo model (see \cite{Fitzhugh1961}) consists of only two equations and is formulated in dimensionless variables, meaning that all state variables lack physical units. This feature is particularly relevant in the present context, since the dimensionality and physical units of the input data provided to the network may differ substantially from those associated with the underlying biophysical model, unless the latter is dimensionless.

After selecting a dimensionless dynamical model for the hidden layer, it becomes necessary to preprocess the input data so as to transform them into dimensionless quantities compatible with the chosen system.

The FitzHugh--Nagumo model is given by
\begin{equation}
\begin{cases}
\dfrac{dv}{dt} = I - v(v - a)(v - 1) - w, \\[0.2cm]
\dfrac{dw}{dt} = b(v - gw),
\end{cases}
\end{equation}
where $I, g \geq 0$, $b > 0$, and $0 < a < 1$ are model parameters. Within this framework, the variable $v$ can be interpreted as the membrane potential, whereas the variable $w$ does not admit a direct physiological interpretation and is introduced to model recovery effects that restore the system toward its resting state.

\begin{figure}[H]
    \centering
    \includegraphics[width=0.6\linewidth]{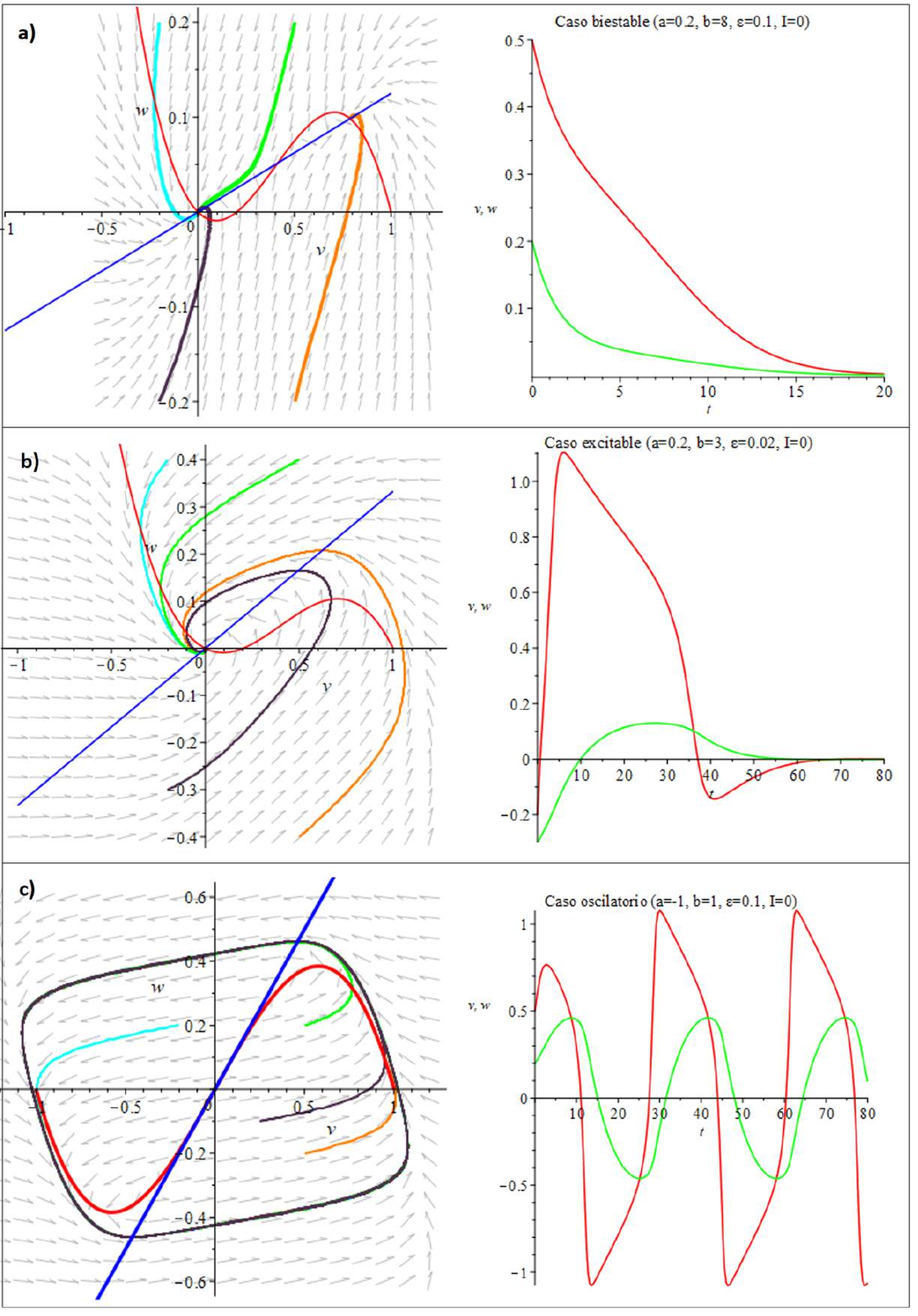}
    \caption{Dynamics of the variable $v$ for different parameter choices \cite{martinez_fitzhugh}. In the phase plane, the nullcline of the variable $v$ is shown as a red cubic curve, while the nullcline of the recovery variable $w$ is represented by a green straight line. Trajectories illustrate the joint evolution of the system from different initial conditions. In the time-domain representations, the membrane potential $v(t)$ is depicted in red and the recovery variable $w(t)$ in green.}
    \label{MemPo1} 
\end{figure}

As illustrated in Figure~\ref{MemPo1}, the FitzHugh--Nagumo system exhibits a rich variety of dynamical behaviors depending on the choice of parameters. For this reason, in our PMNN construction we regard the parameters of the FitzHugh--Nagumo model as hyperparameters of the network, allowing the hidden-layer dynamics to adapt to different regimes.

\section{Numerical experiments with PMNNs}

In their current formulation, PMNNs are more naturally aligned with static prediction tasks than with time-series modeling. This is because the notion of ``time'' appearing in the architecture corresponds primarily to an internal layer-evolution variable induced by an ODE flow, rather than to the physical time index associated with the observed data. In particular, the input specifies an initial condition, and the forward pass integrates the corresponding state up to a chosen internal time, which effectively acts as a continuous-depth transformation.

When applied to temporal data, this design typically reinitializes the dynamics on a per-sample (or per-window) basis and does not inherently maintain a persistent latent state that propagates across successive observations. Such persistence is essential for capturing long-range temporal dependencies. This does not imply that PMNNs are unsuitable for time-series applications, but rather that additional architectural or modeling components are required in order to fully exploit their continuous-time structure in genuinely sequential settings.

In line with these considerations, our experimental study focuses on a static regression setting. In preliminary attempts to benchmark the proposed architecture against continuous-time baselines—namely Neural Ordinary Differential Equations (NODEs) and Closed-form Continuous-time Networks (CfCs)—we removed any explicit temporal dimension from the input data and treated each sample as a static feature vector. This choice allows for a controlled and fair comparison between models.

A detailed pseudocode description of the training procedure for the proposed PMNN architecture is provided below. It explicitly summarizes both the numerical integration routine used to evolve the hidden-layer dynamics and the parameter update steps employed during learning. All model definitions and training scripts are publicly available in the accompanying repository (see \cite{sosa_physics_2025}).

\begin{algorithm}[H]
    \caption{Training of FitzHugh--Nagumo-based PMNN}
    \begin{algorithmic}
        \State \textbf{Input:} Training data $(x_i, y_i)_{i=1}^{N}$, FitzHugh--Nagumo parameters $(a, b, g, I)$
        \State \textbf{Hyperparameters:} Number of Euler steps $H$ (equivalently via $\Delta t$ and $t_{\text{end}}$), learning rate $\alpha$, number of epochs $E$
        \State \textbf{Initialize:} $\theta \gets \{W_h, b_h, W_{\text{out}}, b_{\text{out}}\}$

        \For{each epoch $e = 1, \dots, E$}
            \For{each training sample $(x_i, y_i)$}
                \State $h_0 \gets \phi(W_h x_i + b_h)$ \Comment{LayerNorm + SiLU in our implementation}
                \State $[v, w] \gets h_0$ \Comment{Initial state of the dynamical hidden layer}

                \For{$k = 1, \dots, H$} \Comment{Explicit Euler integration}
                    \State $\dot{v} \gets I - v (v - a)(v - 1) - w$
                    \State $\dot{w} \gets b (v - g w)$
                    \State $v \gets v + \Delta t \, \dot{v}$
                    \State $w \gets w + \Delta t \, \dot{w}$
                \EndFor

                \State $h \gets [v, w]$ \Comment{Terminal hidden state}
                \State $\hat{y}_i \gets W_{\text{out}} h + b_{\text{out}}$ \Comment{Output prediction}
                \State $\ell \gets \mathcal{L}(\hat{y}_i, y_i)$
                \State Compute gradients $\nabla_{\theta}\ell$ using automatic differentiation
                \State Update parameters: $\theta \gets \theta - \alpha \, \nabla_{\theta}\ell$
            \EndFor
        \EndFor
    \end{algorithmic}
\end{algorithm}

\subsection{Implementation details}
The experiments were conducted on a desktop computer equipped with 32 GB of RAM and an Intel Core i7-14700F processor running at 2.1 GHz. The operating system was Windows 11 Home, and the programming environment was Python version 3.12.3. The PyTorch framework \cite{paszke2019pytorch} was used, running on a single NVIDIA GeForce RTX 4070 SUPER GPU.

\subsection{California Housing dataset}

The California Housing dataset was derived from the 1990 U.S. Census and consists of one observation per census block group. A block group is the smallest geographical unit for which the U.S. Census Bureau publishes sample data; it typically contains between 600 and 3,000 people. A household is defined as a group of individuals residing within the same dwelling.

The dataset contains 20,640 instances and includes eight continuous features (such as the median house age within a block group and the average number of bedrooms per household), along with a numerical target variable. The target represents the median house value for California districts, expressed in units of one hundred thousand U.S. dollars (\$100,000).

The dataset was originally obtained from the StatLib repository \cite{pace1997sparse} and can be readily loaded using the Scikit-learn library \cite{scikit-learn}, which is the approach adopted in this study.

To ensure robust model training and evaluation, the dataset was partitioned into three subsets: 70\% for training, 20\% for validation, and 10\% for testing. This split was chosen to balance the availability of data for model learning while retaining a sufficient portion for performance assessment.

Furthermore, all features were standardized using z-score normalization. This preprocessing step is essential when employing learning algorithms that are sensitive to feature scaling, such as gradient descent-based methods. Standardization ensures that each feature contributes proportionally to the learning process, prevents features with larger numerical ranges from dominating the optimization, and typically improves convergence speed and numerical stability during training.

\subsection{Comparison between models}

The FitzHugh--Nagumo-based PMNN introduces a set of hyperparameters intrinsic to the dynamical hidden layer, namely the integration step size $\Delta t$, the integration start time $t_{\mathrm{start}}$, and the terminal integration time $t_{\mathrm{end}}$. These choices determine the effective number of explicit Euler updates,
\[
H = \frac{t_{\mathrm{end}} - t_{\mathrm{start}}}{\Delta t},
\]
and therefore control both the computational cost of the model and the extent to which the hidden state evolves along the FitzHugh--Nagumo flow.

As an initial model-selection strategy, we conducted a manual grid search over plausible values of $\Delta t$, $t_{\mathrm{end}}$, batch size, and learning rate. Throughout these experiments, the integration start time was fixed to $t_{\mathrm{start}} = 0$, in accordance with our implementation, where the dynamical hidden state is initialized directly from an input-dependent mapping. Consequently, several additional hyperparameters were not optimized at this stage, including $t_{\mathrm{start}}$ itself, optimizer-related settings (such as the Adam coefficients $\beta_1$ and $\beta_2$ and weight decay), architectural choices (e.g., the use and type of normalization layers and pre-dynamical nonlinearities), as well as the FitzHugh--Nagumo parameters, which were treated as fixed throughout the reported experiments.

We benchmark the FitzHugh--Nagumo-based PMNN against CfC and NODE baselines. Both baselines were tuned using a grid search under the same selection criterion, namely the mean squared error (MSE). The explored hyperparameter ranges and the corresponding best-performing configurations are summarized in Table~\ref{tab:gridsearch_all_models}.

Since the proposed PMNN is a novel architecture, there is currently no established set of standard hyperparameter values in the literature that can be directly transferred to our setting. For the CfC model, the explored hyperparameter ranges were guided by the original public implementation \cite{cfcgithub} provided by the authors, in particular the configurations reported for traffic prediction tasks. Similarly, the NODE baseline was configured following the original public implementation \cite{torchdiffeq} commonly used for time-series prediction.

\begin{table*}[t]
\centering
\caption{Hyperparameter grid search spaces and selected optimal configurations for the FitzHugh--Nagumo-based PMNN and the baseline models CfC and NODE.}
\label{tab:gridsearch_all_models}
\setlength{\tabcolsep}{4pt} 
\renewcommand{\arraystretch}{1.1}
\begin{tabular}{l l p{5.0cm} l}
\toprule
\textbf{Model} & \textbf{Hyperparameter} & \textbf{Search space} & \textbf{Optimal value} \\
\midrule
\multirow{4}{*}{PMNN}
& Time step
& $\{1,2,5,10,20,50,100,500\}$
& 20 \\
& End time
& $\{10,20,50,100,200,500,$\\
& & $\phantom{\{}1000,10000\}$
& 500 \\
& Batch size
& $\{32,64,128,256\}$
& 32 \\
& Learning rate
& $\{10^{-5},5{\times}10^{-5},10^{-4},5{\times}10^{-4},$\\
& & $\phantom{\{}10^{-3},5{\times}10^{-3},10^{-2},5{\times}10^{-2},$\\
& & $\phantom{\{}10^{-1}\}$
& $5{\times}10^{-4}$ \\
\midrule
\multirow{3}{*}{CfC}
& Number of units
& $\{64,128,256\}$
& 64 \\
& Batch size
& $\{32,64,128,256\}$
& 256 \\
& Learning rate
& $\{10^{-5},5{\times}10^{-5},10^{-4},5{\times}10^{-4},$\\
& & $\phantom{\{}10^{-3},5{\times}10^{-3},10^{-2},5{\times}10^{-2},$\\
& & $\phantom{\{}10^{-1}\}$
& $5{\times}10^{-3}$ \\
\midrule
\multirow{4}{*}{NODE}
& Number of layers
& $\{2,3\}$
& 2 \\
& Hidden dimension
& $\{5,10,15,20\}$
& 15 \\
& Batch size
& $\{32,64,128,256\}$
& 32 \\
& Learning rate
& $\{10^{-4},5{\times}10^{-4},10^{-3},$\\
& & $\phantom{\{}5{\times}10^{-3}\}$
& $10^{-3}$ \\
\bottomrule
\end{tabular}
\end{table*}

Due to computational constraints, we did not perform an exhaustive search over training-related design choices such as the optimizer family or the loss function. Optimization was carried out using Adam with default coefficients, while the learning rate was selected via the grid search described above. Training was run for up to 100 epochs and employed early stopping with a patience of 10 epochs and a minimum improvement threshold (minimum delta) of $5\times 10^{-2}$ in validation loss, in order to reduce overfitting and avoid unnecessary computation.

While more sophisticated hyperparameter optimization methods could potentially yield additional performance gains, we deliberately opted for an initial grid search in order to obtain an interpretable and systematic first characterization of the behavior of this new architecture. Model sizes, reported as the total number of trainable parameters, are shown in Table~\ref{tab:model_parameter_count}.

\begin{table}[ht]
\centering
\caption{Comparison of model size, test performance (RMSE), and total simulation time for the best hyperparameter configuration.}
\label{tab:model_parameter_count}
\begin{tabular}{l r c c}
\toprule
\textbf{Model} & \textbf{\#params} & \textbf{RMSE} & \textbf{Time} \\
\midrule
PMNN & 25    & 0.66 & 54 s \\
CfC  & 14053 & 0.58 & 3 s \\
NODE & 678   & 0.58 & 11 min 20 s \\
\bottomrule
\end{tabular}
\end{table}

\begin{figure}[H]
    \centering
    \includegraphics[width=0.8\linewidth]{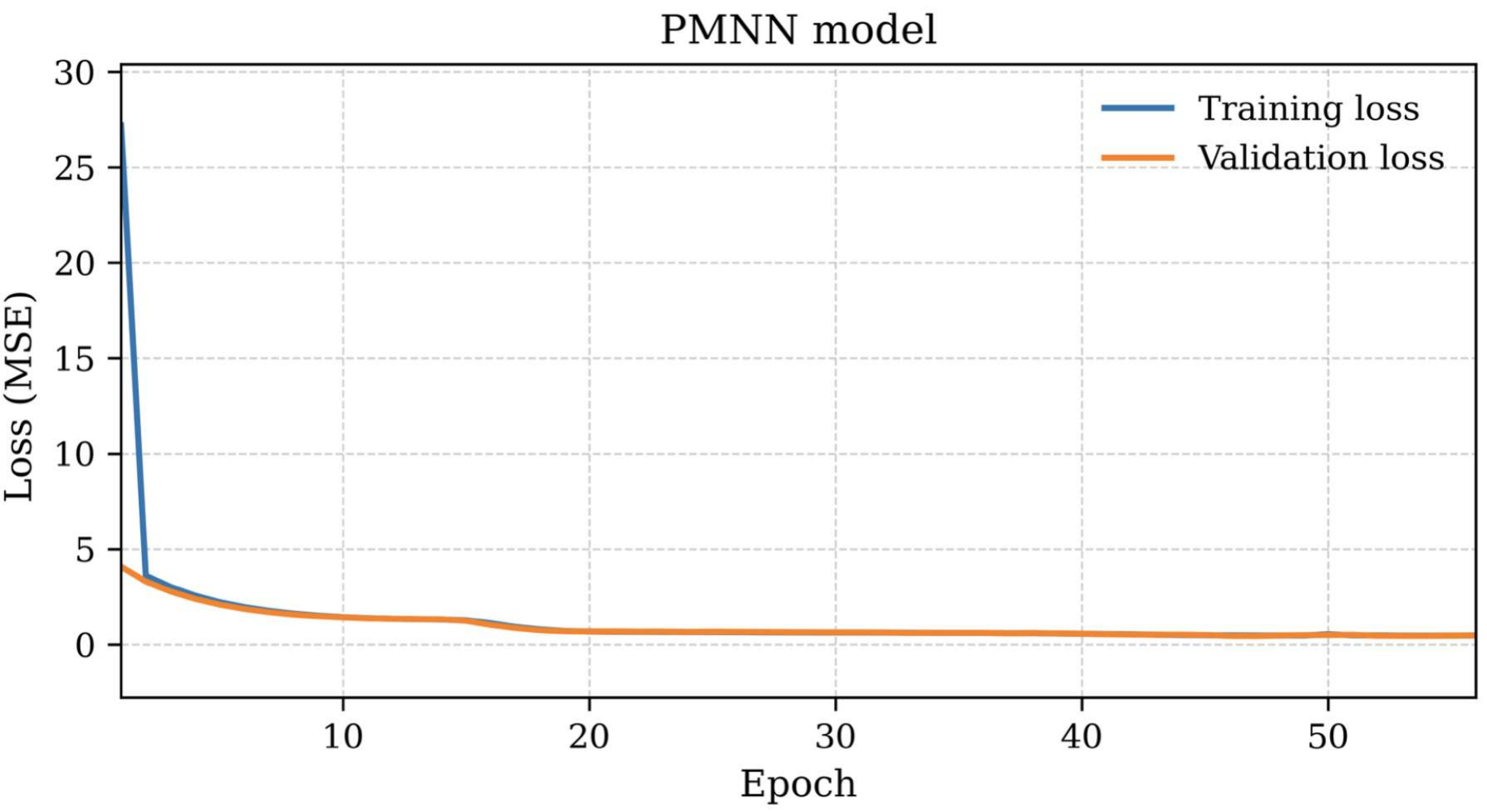}
    \caption{Training and validation loss curves of the PMNN model.}
    \label{train_PMNN} 
\end{figure}

The learning dynamics of the three models further clarify these differences in representational structure. As shown in Figure~\ref{train_PMNN}, the PMNN exhibits a smooth and nearly monotonic decrease in training loss, accompanied by a validation curve that closely tracks the training trajectory throughout the entire optimization horizon. This behavior is consistent with the highly constrained parameterization of the model. The dynamical hidden layer imposes a strong inductive bias that mitigates overfitting even in the presence of limited data, while the small number of degrees of freedom limits the model’s ability to reduce the loss beyond a certain threshold. The gap between training and validation loss remains minimal, indicating stable generalization, but it also reveals that expressive capacity is ultimately bounded by the fixed dynamical structure imposed by the FitzHugh--Nagumo flow.

\begin{figure}[H]
    \centering
    \includegraphics[width=0.8\linewidth]{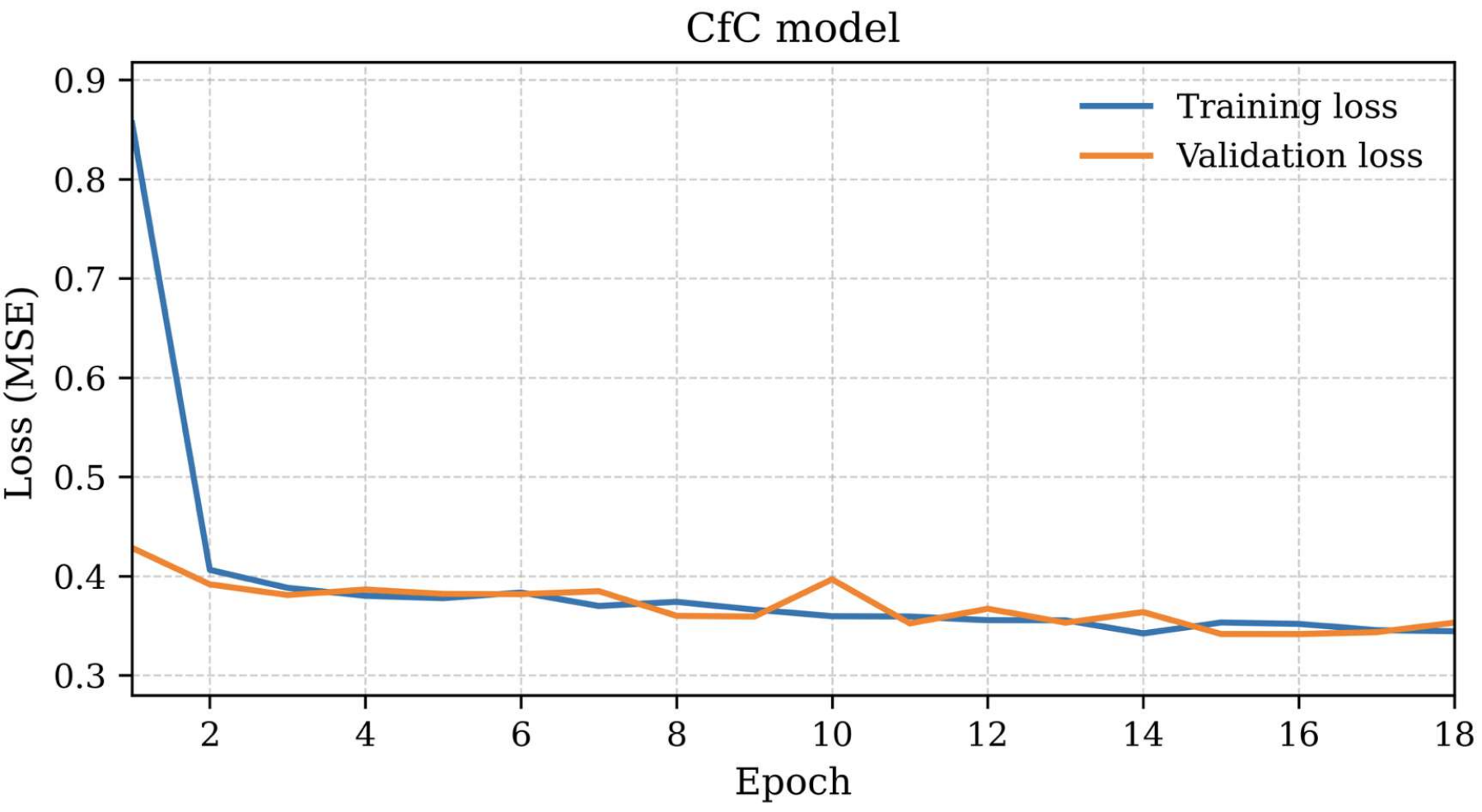}
    \caption{Training and validation loss curves of the CfC model.}
    \label{train_CfC} 
\end{figure}

In contrast, the CfC model (see Figure~\ref{train_CfC}) displays a markedly different optimization profile. The mild divergence between the training and validation curves toward the end of training suggests that the CfC architecture requires careful regularization or early stopping to avoid entering an overfitting regime.

\begin{figure}[H]
    \centering
    \includegraphics[width=0.8\linewidth]{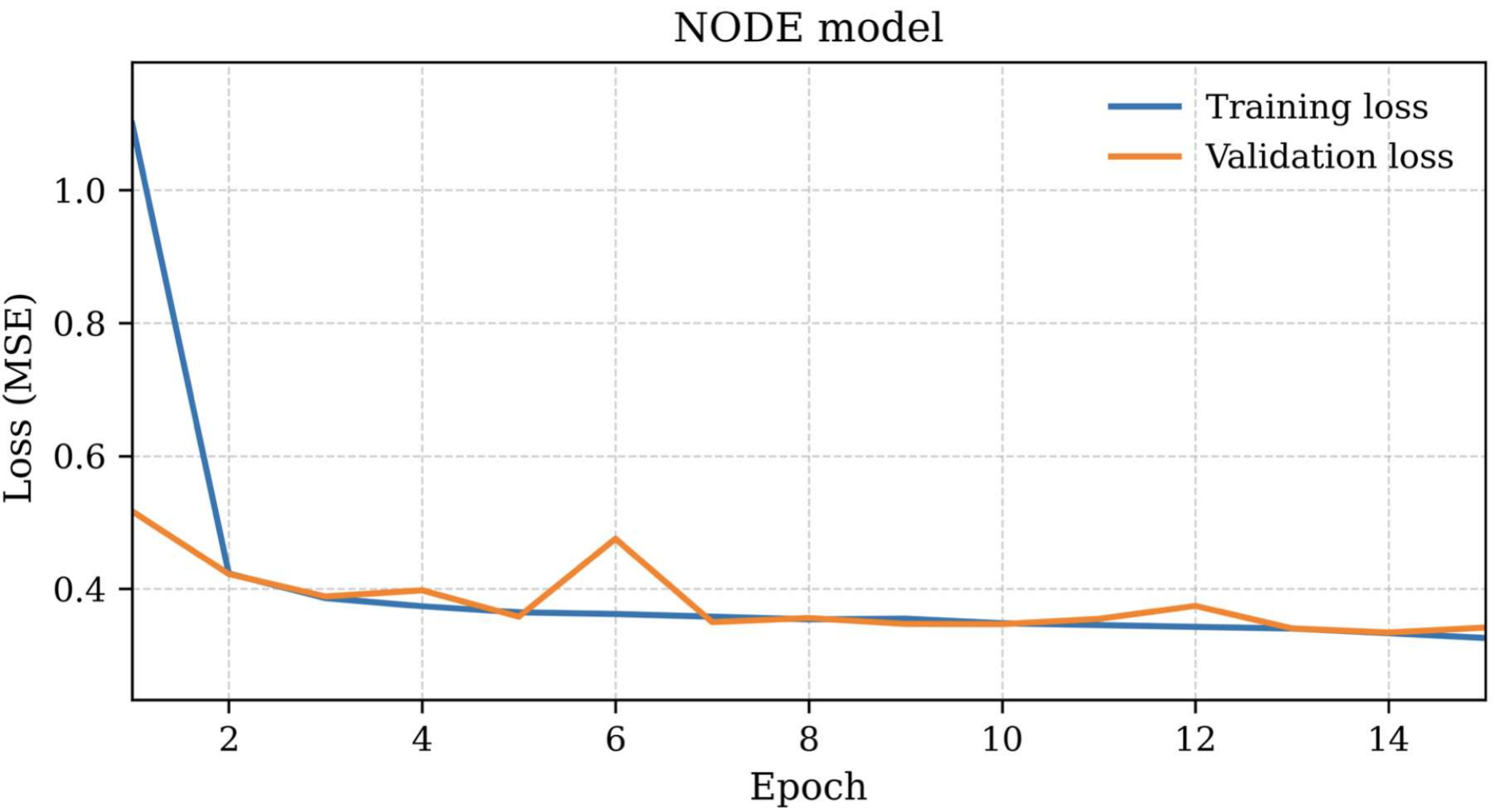}
    \caption{Training and validation loss curves of the NODE model.}
    \label{train_NODE} 
\end{figure}

A similar trend is observed for the NODE baseline (see Figure~\ref{train_NODE}), whose validation loss decreases more rapidly than that of the PMNN but exhibits larger fluctuations, particularly during the early stages of training. This oscillatory behavior reflects the sensitivity of neural ODE solvers to numerical stiffness induced by learned vector fields, as well as the higher complexity of the parameter space they inhabit. While the NODE ultimately attains competitive validation performance, the instability of the training dynamics highlights the additional optimization challenges inherent to unconstrained continuous-time models, in contrast to the strongly regularized behavior enforced by the PMNN’s fixed mechanistic dynamics.

Overall, these results indicate that the proposed PMNN achieves competitive predictive performance despite an extremely small number of trainable parameters. This highlights an interesting trade-off between sample efficiency, model compactness, and the computational overhead induced by internal dynamical simulation. At the same time, the baseline models attain lower test error at substantially larger parameter counts, suggesting that, in the present experimental configuration, the main comparative advantage of the PMNN lies in compactness, interpretability, and a strong inductive bias rather than in absolute predictive accuracy.

\section{Conclusions}

In this work, we have introduced \emph{Dynamical Physics-Modeled Neural Networks} (DynPMNNs), a novel class of deep learning architectures in which the transformations performed by hidden layers are defined as solution operators of ordinary differential equations. In contrast to classical feed-forward neural networks, where nonlinear activation functions are static and memoryless, DynPMNNs endow each layer with an internal state that evolves according to a prescribed dynamical system, whose initial condition is determined by the incoming signal. This construction provides a principled and interpretable mechanism for embedding mechanistic, physics-inspired models directly into the architecture of a neural network.

From a theoretical standpoint, the proposed framework is rigorously grounded in the theory of Reproducing Kernel Banach Spaces (RKBSs). By exploiting representation results for solutions of abstract training problems in RKBSs, we have shown that physics-modeled neural architectures can be interpreted as finite-dimensional realizations arising from infinite-dimensional variational formulations. This perspective clarifies the relationship between classical fully connected networks and the proposed DynPMNNs, and highlights how replacing static activation functions with solution maps of initial value problems leads to a genuine generalization of standard neural architectures.

On the computational side, we instantiated the DynPMNN framework using biophysical neuron models, with particular emphasis on the FitzHugh--Nagumo system. The resulting architecture integrates an explicit Euler scheme into the forward pass, yielding an Euler block whose depth corresponds to the number of integration steps and whose structure naturally induces residual connections. Numerical experiments on the California Housing regression benchmark demonstrate that, despite having an extremely small number of trainable parameters, the proposed DynPMNN achieves competitive predictive performance when compared with Neural Ordinary Differential Equations (NODEs) and Closed-form Continuous-time Networks (CfCs). These results highlight a clear trade-off between interpretability, model compactness, and computational cost: while baseline models achieve slightly lower test error at the expense of substantially larger parameter counts or increased solver complexity, DynPMNNs offer a highly constrained yet expressive alternative driven by strong inductive biases encoded in the underlying dynamics.

Beyond these empirical considerations, the scope of the proposed framework extends beyond the design of more efficient continuous-time neural architectures. In particular, DynPMNNs provide a modeling paradigm that is conceptually aligned with the way biological neural systems process information. Sensory inputs arriving through different modalities may be viewed as external signals that initialize and modulate internal physiological processes, whose evolution unfolds over time according to underlying biophysical mechanisms. From this perspective, modeling hidden layers as dynamical systems governed by differential equations offers a mathematically consistent way to represent how incoming information can activate internal brain processes of a physiological nature, rather than being transformed instantaneously by static nonlinearities.

Finally, the present work opens several directions for future research. From a mathematical perspective, important open questions concern the expressivity, approximation properties, and stability of DynPMNNs as functions of the chosen dynamical system and numerical integration scheme. A systematic analysis of solver-induced biases, step-size selection, and their interaction with learning dynamics is also required. From a modeling standpoint, the framework naturally extends to other classes of differential equations, including higher-dimensional systems, stiff dynamics, and potentially partial differential equations, which would allow the incorporation of spatial structure and geometry into neural architectures. In this sense, DynPMNNs may serve as a bridge between data-driven learning, mechanistic modeling, and physiologically inspired representations of information processing.

\section*{Acknowledgments}
A. M. R. acknowledges support from the Consejería de Educación of the Junta de Castilla y León (Spain) under grant SA062G24.

M. F. C. gratefully acknowledges the financial support provided by the Department of Education of the Junta de Castilla y León and the FEDER Funds (Escalera de Excelencia CLU-2023-1-02).

\bibliographystyle{elsarticle-num}
\bibliography{references}

\end{document}